\documentclass[sigconf]{aamas} 

\usepackage{balance} 
\usepackage{algorithm,algorithmic,amsmath}
\usepackage{url}
\usepackage{hyperref}
\usepackage{multirow}

\setcopyright{none}
\acmConference[ALA '23]{Proc.\@ of the Adaptive and Learning Agents Workshop (ALA 2023)}
{May 9-10, 2023}{Online, \url{https://ala2023.github.io/}}{Cruz, Hayes, Wang, Yates (eds.)}
\copyrightyear{2023}
\acmYear{2023}
\acmDOI{}
\acmPrice{}
\acmISBN{}
\acmSubmissionID{???}

\title[AAMAS-2023 Formatting Instructions]{ADT: Agent-based Dynamic Thresholding for Anomaly Detection}

\author{Xue Yang}
\affiliation{
  \institution{University of Galway}
  \city{Galway}
  \country{Ireland}}
\email{x.yang6@universityofgalway.ie}

\author{Enda Howley}
\affiliation{
  \institution{University of Galway}
  \city{Galway}
  \country{Ireland}}
\email{enda.howley@universityofgalway.ie}

\author{Michael Schukat}
\affiliation{
  \institution{University of Galway}
  \city{Galway}
  \country{Ireland}}
\email{michael.schukat@universityofgalway.ie}

\begin{abstract}
The complexity and scale of IT systems are increasing dramatically, posing many challenges to real-world anomaly detection. Deep learning anomaly detection has emerged, aiming at feature learning and anomaly scoring, which has gained tremendous success. However, little work has been done on the thresholding problem despite it being a critical factor for the effectiveness of anomaly detection. In this paper, we model thresholding in anomaly detection as a Markov Decision Process and propose an agent-based dynamic thresholding (ADT) framework based on a deep Q-network. The proposed method can be integrated into many systems that require dynamic thresholding. An auto-encoder is utilized in this study to obtain feature representations and produce anomaly scores for complex input data. ADT can adjust thresholds adaptively by utilizing the anomaly scores from the auto-encoder and significantly improve anomaly detection performance. The properties of ADT are studied through experiments on three real-world datasets and compared with benchmarks, hence demonstrating its thresholding capability, data-efficient learning, stability, and robustness. Our study validates the effectiveness of reinforcement learning in optimal thresholding control in anomaly detection.
\end{abstract}


\newcommand{\BibTeX}{\rm B\kern-.05em{\sc i\kern-.025em b}\kern-.08em\TeX}

\begin{document}


\pagestyle{fancy}
\fancyhead{}


\maketitle 


\section{Introduction}

Anomaly detection has attracted much attention for years. It refers to the problem of finding patterns in data that deviate from the expected normal behavior \cite{chalapathy2019deep}. Anomalies are often "rare" and "different" and can provide critical information ~\cite{liu2012isolation}. Anomaly detection has been well studied in numerous research areas and application domains including data mining, network intrusion detection, medical and public health, and energy management ~\cite{chandola2009anomaly}. 

In general, a measurement of the input data (i.e. anomaly score) will be compared with a domain-specific threshold to determine the degree to which the data is considered abnormal. There is a substantial amount of research that is dedicated to using deep learning (DL)-based methods for feature learning and anomaly scoring for complex input data, achieving competitive advantages over conventional anomaly detection methods. However, little work has been done on the thresholding problem despite it being a critical factor for the effectiveness of anomaly detection. Given the fact that varying the threshold values can greatly change detection performance, the importance of thresholding should receive more attention.

In the literature, many existing anomaly detection approaches use static or expert-defined thresholds ~\cite{audibert2020usad, lin2020anomaly, niu2020lstm}. However, with the dramatic increase in the complexity and scale of IT systems and data, the traditional thresholding approaches are not efficient enough due to their poor scalability and robustness ~\cite{audibert2020usad}. They have difficulty adapting to non-stationary and evolving time series data ~\cite{chalapathy2019deep}. Recently, some novel dynamic thresholding methods were proposed such as the statistical method that is based on extreme value theory (EVT) ~\cite{siffer2017anomaly}. Unfortunately, some of the methods cannot always guarantee good results and the improvement over static thresholding is limited, which has also been shown in our experiments. It is desirable to develop more effective thresholding methods in this area.

Reinforcement Learning (RL) is well known for its capability of solving complex sequential decision-making in various domains, e.g. computer games, robotics, transport, and traffic signal control. RL models the sequential decision-making problem as a Markov Decision Process (MDP), where an agent learns through trial and error from the interaction with an environment. The goal for the agent is to maximize the total rewards over time. An RL agent is capable of providing advantageous control for dynamic complex systems ~\cite{kuhnle2021designing}. The application of RL in optimal thresholding control is therefore a worthwhile study. 

This paper presents an agent-based dynamic thresholding (ADT) framework for optimal thresholding control in anomaly detection and models the problem as a MDP. The popular deep reinforcement learning (DRL) algorithm, i.e. Deep Q-network (DQN) is applied to deal with high-dimensional and continuous inputs. An auto-encoder (AE) is utilized to obtain feature representations and perform anomaly scoring for time series data. ADT can provide appropriate dynamic thresholds by utilizing the anomaly scores from AE, achieving adaptive anomaly detection. Specifically, the threshold $\delta$ is adjusted between a passive mode (i.e. $\delta=1$) and an active mode (i.e. $\delta=0$) in the detection process. 

The proposed thresholding method can be combined with many systems that require dynamic thresholding, ideally as long as they are able to generate expressive abnormal-related measurements. ADT is lightweight and data-efficient, since a small amount of data is adequate for the training (e.g. $<1\%$ of the benchmarked dataset). The experimental results on three real-world datasets prove that our thresholding method leads to significantly improved detection performance compared with benchmarks, e.g. achieving the best F1 scores in all cases ranging from 0.945 to 0.999. 

The main contributions of this paper are as follows: first of all, we model thresholding in anomaly detection as a MDP and propose an agent-based dynamic thresholding framework using DQN. Secondly, we incorporate a deep generative model and our agent-based thresholding controller to perform anomaly detection on real-world datasets, demonstrating the high detection performance, data efficiency, stability, and robustness. Finally, we conduct a feasibility study to further validate the thresholding performance of the proposed method and demonstrate its superiority over benchmarks.

In the rest of this paper, we begin with a literature review of related work and background knowledge in Section \ref{sec:related work}. We then introduce the details of our methods in Section \ref{sec:method}, and present the experiments and results in Section \ref{sec:Experimental Setup} and Section \ref{sec:experiments and results}. Finally, we conclude the paper in Section \ref{sec:conclusion}.


\section{Background and Related Work}
\label{sec:related work}
This section introduces the background knowledge and related work of unsupervised deep anomaly detection, dynamic thresholding in anomaly detection, and reinforcement learning.
\subsection{Unsupervised Deep Anomaly Detection} 
\label{sec:unsupervised anomaly detection}
 In the past, commonly-used anomaly detection methods included classification-based methods such as SVM \cite{tax2004support,ma2003time,amer2013enhancing}, clustering-based methods such as K-means \cite{li2010research,zhang2009mixed}, and density-based methods such as KNN \cite{chaovalitwongse2007time,su2011real}. However,  such traditional methods generally suffer from the curse of dimensionality of high-dimensional inputs, and are limited in capturing complex temporal information in time series \cite{chalapathy2019deep,choi2021deep}. Unbalanced training data with very few available labels is also an issue \cite{lin2020anomaly}. Recent studies have shown that unsupervised deep anomaly detection (UDAD) helps to address these gaps.

There are various UDAD approaches based on variational auto-encoder (VAE), recurrent neural networks (RNN), generative adversarial networks (GAN), and other deep neural network architectures or their variants. The VAE-LSTM \cite{lin2020anomaly} employs VAE to extract local information of short windows and LSTM to estimate long-term sequence correlations, which can detect anomalies over both short and long periods. USAD \cite{audibert2020usad} embeds two AEs within an adversarial training framework to combine the advantages of both techniques while mitigating the limitations of each. The LSTM-based VAE-GAN \cite{niu2020lstm} jointly trains the LSTM-based encoder, generator, and discriminator to take advantage of the mapping ability and the discrimination ability simultaneously. The deep auto-encoding Gaussian mixture model (DAGMM) \cite{zong2018deep} utilizes AE to obtain low-dimensional features and reconstruction errors, and feed them to a Gaussian mixture modeling to perform density estimation. 

The aforementioned UDAD methods focus on generic feature extraction, representation learning of normality, and anomaly score learning \cite{pang2021deep}, which have gained tremendous success in this area. However, some of the studies did not clearly explain their thresholding approach \cite{audibert2020usad}, and some used static or expert-defined thresholds \cite{lin2020anomaly, zong2018deep, niu2020lstm}. In fact, the thresholding problem is critical in anomaly detection as it highly affects the detection results. The underestimation of thresholding can result in a bottleneck of an anomaly detection method and therefore is worth further research.   

\subsection{Dynamic Thresholding in Anomaly Detection}
\label{sec:dynamic thresholding in anomaly detection}
Dynamic thresholding adjusts thresholds in a timely manner to adapt to complex and dynamic scenarios. It is a promising technique that may help to achieve surpassed anomaly detection performance over static or manually-defined thresholds.

In recent years, some dynamic thresholding approaches have been proposed in the anomaly detection domain. \cite{siffer2017anomaly} proposed a statistical method based on EVT and introduced two algorithms for thresholding in stationary and drifting contexts, respectively. \cite{hundman2018detecting} presented an unsupervised and non-parametric thresholding approach for spacecraft anomaly detection without statistical assumptions about prediction errors. \cite{xie2021graph} proposed a thresholding method by analyzing periodic time windows and using the moving weighted average to calculate thresholds. 

RL has been applied for dynamic thresholding in anomaly detection lately. \cite{yu2020policy} proposed a policy-based anomaly detector PTAD which adopted the DRL algorithm, i.e. asynchronous advantage actor-critic (A3C), and embedded it with LSTM components. Their method enables the threshold to be adjusted for the trade-off between precision and recall. \cite{watts2021dynamic} modeled anomaly detection in the transportation domain as a partial observable MDP and used convolutional neural network (CNN) and A3C to solve it, which can generate dynamic classification thresholds. 

RL has been successfully employed to provide adaptive optimal control \cite{chen2019reinforcement} in various domains. The related background knowledge of RL is introduced in the next section.

\subsection{Reinforcement Learning}
\label{sec:DQN}
RL is a machine learning paradigm where an agent learns an optimal policy by trial and error through interacting with an environment. RL models the sequential decision-making problem as a MDP, which defines the interactions between a learning agent and an environment in the mathematical formalization \cite{sutton2018reinforcement}.

MDP is described as a tuple of $(S,A,T,R,\gamma)$ that includes a set of states $S$, a set of actions $A$, a state transition function $T$, a reward function $R$, and a discount factor $\gamma$. At each time step {$t \in \{{1,2,3,...}$\}}, the agent perceives an environment state $s_t$ and selects an action $a_t$ following a policy $\pi(a_t|s_t)$, i.e. a mapping from state $s_t$ to action $a_t$. Correspondingly, the environment will respond to the agent's action and send to the agent a reward $r_{t}$ and a new state $s_{t+1}$. The state $s_t$ should have the Markov property and convey substantial information about the history that affects the future \cite{sutton2018reinforcement}. The value of a state-action pair is denoted as 
 $q_\pi(s,a)$, which is the expectation of the discounted sum of long-term rewards after following $\pi$ in state $s$ and taking action $a$. It is denoted as:
\begin{equation}
    \label{eqn:value function}
    q_{\pi}(s,a)=\mathbb{E}_{\pi}[\sum_{k=0}^{\infty} \gamma^{k}R_{t+k}|S_t=s, A_t=a]
\end{equation}

The objective of the agent is to find the optimal policy $\pi^*$ with the optimal state-action value.

In this study, we apply a popular DRL algorithm, i.e. DQN \cite{mnih2015human} to solve the MDP.  DQN is based on temporal-difference Q-learning aiming to approximate the state-action value function $Q(s,a;\theta)$ with a neural network called the Q-network, where $\theta$ denotes the network weights. It incorporates experience replay and a separate target network (i.e. $\hat{Q}$ with weights $\theta^{-}$) to mitigate correlations between training samples and achieve stable training. DQN is able to handle continuous inputs, which is suitable for our method since the anomaly scores used as part of the ADT inputs are continuous values.

According to \cite{mnih2015human}, the Q-learning update in the training process follows Equation \ref{eqn:Q-learning update} with a random minibatch of transitions $(s_j,a_j,r_{j},s_{j+1})$ sampled from the replay memory $D$:

\begin{equation}
\label{eqn:Q-learning update}
 y_j=
 \begin{cases} 
    r_j,                                                     & \text{if terminal} \\
    r_j+{\gamma\ max}_{a^{'}}\hat{Q}(s_{j+1},a';\theta^{-}), & \text{otherwise }
    \end{cases}
\end{equation}
where $y_j$ is the target $Q$ value that is generated based on the target-network $\hat{Q}$. The gradient descent is performed on the loss between the predicted $Q(s_j,a_j;\theta)$ and the target $y_j$ to update the weights of the Q-network.
 

\section{Method}
\label{sec:method}
In this section, the problem we are addressing is first formalized in Section \ref{sec:problem formulation}. The unsupervised DL method for feature learning and anomaly scoring is introduced in Section \ref{sec:unsupervised ae}. The core of our method, i.e. the agent-based dynamic thresholding framework is presented in Section \ref{sec:adt framework}. Lastly, the method's implementation is described in Section \ref{sec:implementation}.

\subsection{Problem Formulation}
\label{sec:problem formulation}
A time series is a sequence of data points {$X=\{{x_1,x_2,...,x_n}$\}} where $x_i\in R^m$ is an m-dimensional reading at the $i$th time stamp. The data can be normalized between 0 and 1 and split into a sequence of sliding windows with a length $\tau$ and a stride of 1. All the data is converted to a sequence of windows on that basis, i.e. $W=\{{w_1,w_2,...,w_{n-\tau+1}}$\} where $1\leq\tau<n$. We redefine the reading at time step $t$ as the time window {$w_t=\{{x_{t-\tau+1},x_{t-\tau+2},...,x_t}$\}} ($\tau\le t$). The ground truth ${y_t}\in \{{0,1}$\} is the label of $w_t$ where $y_t=1$ indicates an abnormal window and $y_t=0$ indicates a normal window. In anomaly detection with dynamic thresholding, the anomaly score of a time window, i.e. $\mathcal{A}^{scr}_{t}$, is compared with a dynamic threshold $\delta_{t}$ to determine the abnormality of $w_t$, where $\mathcal{A}^{scr}_{t}$ is between 0 and 1. If the anomaly score exceeds the threshold, the window will be considered abnormal. Therefore, the anomaly detection task is to predict a binary label $\hat{y_t}\in \{{0,1}$\} of $w_t$ in the testing dataset.

The detection performance is evaluated by Precision (P), Recall (R), and F1 score (F1):
\begin{eqnarray}
    \label{eqn:evaluation metrics}
    P= \frac{TP}{TP+FP},\  R=\frac{TP}{TP+FN},\ F1=\frac{2\cdot P\cdot R}{P+R}
\end{eqnarray}
where $TP$, $FP$, and $FN$ denote the number of true positives, false positives, and false negatives, respectively.

We use a point-adjust approach proposed by ~\cite{xu2018unsupervised} to label $W$: A window will be deemed as an anomaly whenever there are one or more anomaly points in the sequence; Otherwise, the window will be declared as normal. This is for the consideration that in real-world applications, we are more concerned about the evaluation metrics of contiguous abnormal segments rather than the point-wise metrics ~\cite{xu2018unsupervised}. However, the length of the window should not be too long to ensure the detection performance is not overestimated.

The accuracy of anomaly scoring on unknown data and the suitability of selected thresholds are two important factors that directly determine the anomaly detection performance. In the following, an unsupervised DL method for feature learning with anomaly scoring and a DRL method for dynamic thresholding are introduced. 

\subsection{Unsupervised Feature Learning and Anomaly Scoring}
\label{sec:unsupervised ae}
AE-based anomaly detection methods have been investigated in many studies such as \cite{chalapathy2019deep,choi2021deep,basora2019recent}. AE is a deep generative model that contains an encoder $E$ and a decoder $D$.
The encoder takes an input $X$ and outputs a compressed latent representation $Z$, and the decoder reconstructs the data from $Z$ to $X'$. The difference between $X$ and $X'$ is called the reconstruction error, which serves as the anomaly score $\mathcal{A}^{scr}$. The target of training AE is to minimize the reconstruction error and make $X'$ as close as possible to the original input $X$. The reconstruction error or loss is defined as:
\begin{eqnarray}
    \mathcal{L}_{AE}=||X-AE(X)||_{2}
\end{eqnarray}
where $AE(X)=D(Z)$, $Z=E(X)$, $\mathcal{L}_{AE}$ is the reconstruction error, and ${||.||}_2$ denotes the $L_2$ norm. 

The AE-based anomaly detection method falls under the unsupervised learning category \cite{basora2019recent} and solely normal data is used for training. It enables the learning of feature representations of normal instances and leads to relatively high reconstruction errors over unseen anomalies. This makes it possible to distinguish normal and abnormal instances by comparing the reconstruction error with a threshold. Consequently, the trained AE is able to classify the unknown data into normal or abnormal in the testing stage.
 
Fixed or static thresholds have been applied in many AE-based anomaly detection studies. However, this does not always hold in practice, as some anomalies are close to normal with relatively low reconstruction errors. In this case, a fixed threshold that is capable of separating anomalies with usually high reconstruction errors from normal instances will become infeasible. Moreover, the complexity and non-stationarity of data distribution can change the definition of anomalies implicitly. A static threshold will expire and fail to handle the context drift. Therefore, there is a need for more adaptive thresholding techniques in this area.
 
\subsection{Agent-based Dynamic Thresholding}
\label{sec:adt framework}
We propose an advanced agent-based framework ADT for optimal thresholding control. It aims to maximize anomaly detection performance by issuing proper dynamic thresholds. We model the problem as a MDP and solve it with DQN.

\subsubsection{State}
\label{sec:dqn-state}
The proper definition of state space is essential in MDP. The state space $\mathcal{S}$ should contain substantial information for the agent to make decisions. In this study, the state at time step $t$ consists of six elements, i.e. $s_{t}=\{\mu_{t}, \sigma_{t}, \rho^{TP}_{t}, \rho^{TN}_{t}, \rho^{FP}_{t}, \rho^{FN}_{t}\}$. $\mu_t$ and $\sigma_t$ are the average and variance of the anomaly scores of the encountered previous $k$ windows {$\{{w_{t-k},w_{t-k+1},...,w_{t-1}}$\}}:
\begin{equation}
    \mu_t=\frac{\sum_{t-k}^{t-1} \mathcal{A}^{scr}_{i}}{k}
\end{equation}
\begin{equation}
 \sigma_t=
 \begin{cases}
    0,                                                           & \text{if } k=1\\
    \frac{\sum_{t-k}^{t-1} (\mathcal{A}^{scr}_{i}-\mu_t)^2}{k-1},& \text{otherwise } 
    \end{cases}
\end{equation}
where $t>k\ge1$, $\rho^{TP}_{t}$, $\rho^{TN}_{t}$, $\rho^{FP}_{t}$, and $\rho^{FN}_{t}$ are the percentages of $TP$, $TN$, $FP$, and $FN$ windows out of the $k$ windows.

The objective of this state representation is to capture the dynamics of anomaly scores as well as the effects of previous actions to get more useful context information. By observing the state and interacting with the environment, the agent is able to learn an optimal policy in the MDP to issue an appropriate threshold. 

\subsubsection{Action} 
Given a state $s_{t}$ from the environment, the agent chooses an action {$a_t\in\{{0,1}$\}} which represents the threshold value at time step $t$. The threshold is compared with an anomaly score to determine whether $w_t$ is abnormal. Within the scope of this study, a binary threshold is adequate to achieve a well-performing detection result and is efficient due to its simplicity. 

We define the binary threshold as an active mode ($\delta=0$) and a passive mode ($\delta=1$) given that $\mathcal{A}^{scr}_{t}$ is between $0$--$1$. If a window is abnormal, any threshold value that is lower than its anomaly score leads to a correct determination. The threshold $\delta=0$ is therefore appropriate in this scenario. In other words, it triggers an active mode to be more concerned about anomalies. Conversely, if the window is normal, any threshold value higher than its anomaly score works. The threshold $\delta=1$ hence can guarantee the correctness of the classification and it implies a passive mode to be less concerned about anomalies.

\subsubsection{Reward}
The definition of the reward signal is another essential part of MDP. A straightforward way is to define the reward function by the findings of $TP$, $TN$, $FP$, and $FN$:
\begin{equation}  
\label{eqn:literature reward}
        \begin{cases}
        r_{TP}=r_{TN}=1 \\
        r_{FP}=r_{FN}=-1
        \end{cases}  
\end{equation} 
where the reward for a $TP$ or $TN$ is 1, and the reward for a $FN$ or $FP$ is -1. However, this reward function cannot adapt to different environments and various user requirements. To improve it, we propose a parameterized reward function to assign different importance to $TP$, $TN$, $FP$, and $FN$. Two parameters $\alpha$ and $\beta$ are added to take into account both simplicity and experimental performance:
\begin{equation}
\label{eqn:new reward}
r=\alpha\times(n_{TP}-n_{FP}-n_{FN})+\beta\times n_{TN}\\
\end{equation}
where $n_{TP}$, $n_{TN}$, $n_{FP}$, and $n_{FN}$ denote the numbers of $TP$, $TN$, $FP$, and $FN$ samples found in the observed $k$ time windows of the environment state. The values of $n_{TP}$, $n_{TN}$, $n_{FP}$, and $n_{FN}$ are therefore all integers from $0$ to $k$. $\alpha$ and $\beta$ are non-negative weights that sum up to 1. 

A relatively larger $\alpha$ is supposed to award more to $TPs$, and penalize more to $FPs$ and $FNs$. Different values of $\alpha$ and $\beta$ may twist the policy to result in various performances. The effect of $\alpha$ and $\beta$ on the detection performance is further investigated in our experiments (see Section \ref{sec:effect of parameters}). This parameterized reward design demonstrates its flexibility and effectiveness in our study. It also opens the gate to modeling anomaly detection as a multi-objective problem because in industries, there are many scenarios that the detection system needs to balance between different objectives.

\subsubsection{Training and Inference} In practice, DQN often suffers from training efficiency. We use the following three approaches to mitigate this problem within ADT:
\begin{enumerate}
    \item Due to the fact that both normal and abnormal data are generally contiguous segments, the action, i.e. the threshold is not necessarily to be changed per time step. We make the DQN agent change the action $a$ per $l$ time steps ($l\geq1$) where the exploration and exploitation are balanced; and within the $l$ time steps the agent maintains the action $a$ unchanged. Here is an example with $l=10$: if $t\mathbin{\%}10=0$, the action $a_t$ is from the $\epsilon$-greedy strategy; otherwise, the action is maintained as $a_t=a_{t-1}$. $l=1$ is a special case where the action is changed at every time step. This approach can reduce the overhead of feed-forward procedures in neural networks while holding the stability of the training process. 
    \item A very small amount of data is used for the training of ADT (e.g. less than $1\%$ of the benchmarked dataset), which highly reduces the training time.
    \item The $Q$-network in the DQN is updated only at the end of each training episode to increase the stability of the training process and reduce the computational expense.
\end{enumerate} 

\begin{figure}[h]
    \centering
    \includegraphics[width=0.9\linewidth]{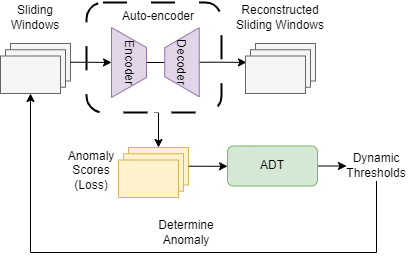}
    \caption{AE-ADT adaptive anomaly detection framework}
    \label{fig:inference}
    \Description{AE-ADT anomaly detection framework. It contains two modules: anomaly score learner and dynamic thresholding controller. The training dataset contains both normal and anomalous.}
\end{figure}

In the inference stage, the trained ADT is able to generate the optimal threshold at each time step to detect anomalies on the testing dataset, and importantly, $l$ should be $1$ in this stage to ensure the learned optimal policy is performed. The training and detection processes are illustrated in Algorithm \ref{alg:DQN training} and Figure ~\ref{fig:inference}.
\begin{algorithm}[htb]
\caption{ADT training algorithm}
\label{alg:DQN training} 
    \begin{algorithmic}
        \STATE Initialize agent and environment
        \STATE Initialize Q-network ($Q$) with random weights $\theta$
        \STATE Initialize target-network ($\hat{Q}$) with weights $\theta^{-}=\theta$
        \STATE Initialize experience replay memory $D$
        \FOR{$episode\in \{{1,2,3,...,M}$\}}
            \FOR{$t\in \{{1,2,3,..., T}$\}}
                \IF{$t\mathbin{\%}l=0$}
                    \STATE Following $\epsilon$-greedy strategy
                    \STATE
                    $\begin{cases}
                        \text{select a random action $a_t$},    & \text{with probability $\epsilon$}\\
                        a_t=argmax_{a}Q(s_t,a;\theta), & \text{otherwise}
                    \end{cases}$
                \ELSE
                    \STATE $a_t=a_{t-1}$
                \ENDIF
            \ENDFOR
            \STATE Execute $a_t$ and get the next state $s_{t+1}$ and reward $r_t$
            \STATE Store transition $(s_t, a_t, r_t, s_{t+1})$ in $D$
            \IF{$t=T$}
                \STATE Sample a random minibatch $(s_j,a_j,r_{j},s_{j+1})$ from $D$
                \STATE $y_j=
                \begin{cases} 
                            r_j,                                                     & \text{if terminal} \\
                            r_j+{\gamma\ max}_{a^{'}}\hat{Q}(s_{j+1},a';\theta^{-}), & \text{otherwise }
                \end{cases}$
                \STATE Perform gradient descent on $(y_j-Q(s_j,a_j;\theta))^2$ with respect to the Q-network parameters
            \ENDIF
            \STATE Decay $\epsilon$
            \STATE Every $C$ episodes, copy weights from $Q$ to $\hat{Q}$
        \ENDFOR 
    \end{algorithmic}
\end{algorithm}

\subsection{Implementation}
\label{sec:implementation}
The workflow of our adaptive anomaly detection method consists of four phases. 
\begin{enumerate}
\item Data pre-processing\\
The time series data is normalized into the range of $0$--$1$ and split into sliding windows of length $\tau$ with stride=1. Note that no down-sampling is used in this phase to avoid the risk of losing information.
\item AE training\\
The AE model is trained using only normal instances. The well-trained AE can perform anomaly scoring for each time window.
\item ADT training\\
ADT is trained with both normal and abnormal instances. The trained ADT can generate the optimal threshold for each time window by utilizing the anomaly score calculated by AE.
\item Evaluation\\
The last phase is the utilization of the trained AE from phase 2 and the trained ADT from phase 3 to do online anomaly detection over the testing dataset. As a time window arrives, if the anomaly score generated by AE is higher than the threshold generated by ADT, the window is considered abnormal.

\end{enumerate}

\section{Experimental Setup}
\label{sec:Experimental Setup}
In this section, we introduce three public datasets, two thresholding methods as benchmarks, and the parameter settings of our method.
 \subsection{Datasets}
 \label{sec:datasets}
 Three real-world datasets were used in our experiments. The characteristics of each dataset and the selection of training sets are described in the following.
 
\textit{Yahoo A1Benchmark} (Yahoo)\footnote{\url{https://webscope.sandbox.yahoo.com/catalog.php?datatype=s&did=70}} is real data representing the metrics of various Yahoo services. It is one of the four benchmarks of a labeled anomaly detection dataset released by Yahoo lab. It has 67 CSV files containing 94866 points in total with an anomaly rate of $1.76\%$. The data is univariate with $m=1$.  

\textit{Secure Water Treatment} (SWaT)\footnote{\url{https://itrust.sutd.edu.sg/itrust-labs_datasets/dataset_info/}} is an operational testbed for a real-world water treatment system producing filtered water, which is widely used in the field of cyber and physical system security \cite{mathur2016swat}. It consists of 11-day continuous operations where 7 days are under normal scenarios and 4 days are under attack scenarios \cite{goh2017dataset}. SWaT is collected from all the 51 sensors and actuators with $m=51$. The dataset contains two CSV files: one with only normal data containing 496800 points, and one with labeled attacks containing 449919 points with an anomaly rate of $11.98\%$.

\textit{Water Distribution} (WADI) is a testbed conducted for the secure water distribution system which is physically connected to the SWaT testbed \cite{ahmed2017wadi}. It consists of 16-day continuous operations where 14 days are under normal scenarios and 2 days are under attack scenarios. WADI is collected from 123 sensors and actuators with $m=123$. Similarly to SWaT, WADI contains two CSV files: one with only normal data containing 1048571 points, and one with labeled attacks containing 172801 points with an anomaly rate of $5.99\%$.

After data pre-processing, AE is trained using only normal time windows, whereas ADT is trained with both normal and abnormal time windows. For ADT, about $0.35\%$ of Yahoo, $0.11\%$ of SWaT, and $0.58\%$ of WADI were used as the training sets, separately, in our study. The training set size for each dataset and the anomaly ratio in the training set were experimentally determined based on anomaly detection performance and data efficiency, however, they can be flexibly adjusted. It is notable that the effect of different values of the anomaly ratio in the training set is beyond the scope of this paper, nevertheless, is worth studying in the future.

\subsection{Benchmarks}
\label{sec:benchmarks}
We use two thresholding approaches as benchmarks to evaluate the performance of ADT in anomaly detection.
\subsubsection{Optimal Static Thresholding Method}
Due to the simplicity and relative effectiveness, the optimal static thresholding (i.e. Static) method has been used as a default thresholding method in many anomaly detection studies such as \cite{audibert2020usad,xu2018unsupervised,lin2020anomaly}. Typically, a value that generates the best performance will be defined as the threshold. For example, if the anomaly scores for the input data are in the range of $a$ to $b$, the optimal threshold will then be selected between $a$ and $b$ with the best F1 score or other metrics. This can be simply implemented through brute force search. 

\subsubsection{EVT-based Dynamic Thresholding Method}
 ~\cite{siffer2017anomaly} proposed a statistical thresholding method which can generate continuous dynamic thresholds. It is based on EVT that aims to fit the distribution of extreme events to a generalized Pareto distribution without a strong hypothesis on the original distribution. Their work relies on the assumption that in any distribution abnormal instances occur with low probability while normal instances occur with high probability. They proposed SPOT and DSPOT algorithms to work in stationary and non-stationary scenarios. DSPOT\footnote{\url{https://github.com/Amossys-team/SPOT}} was used in our study. 

\subsection{Parameter Settings}
\label{sec:parameter settings}
For each dataset, the parameter settings of ADT are displayed in Table \ref{table:parameters}, including the number of training episodes, the time window size $\tau$, the parameter $k$ used in the environment's state representation, the parameter $l$ used in the agent's action selection, and $\alpha$ and $\beta$ for the reward function. These parameters are essential in the reimplementation of our work and the parameter values are experimentally determined. The effects of the important parameters on the performance of ADT are introduced in Section \ref{sec:experiments and results}.
\begin{table}[h]
  \caption{ADT parameters for each dataset}
  \label{table:parameters}
  \begin{tabular}{cccccc}\toprule
    Dataset & Episode & $\tau$ & $k$ & $l$ & $(\alpha,\beta)$ \\ \midrule
    Yahoo &20000   & 10    & 2 & 1  & (0.9, 0.1) \\
   SWaT   &20000 & 12  & 2 & 10  & (0.9, 0.1) \\
   WADI   &20000  & 10   & 2 &  10 & (0.9, 0.1) \\ \bottomrule
  \end{tabular}
\end{table}

\section{Experiments and Results}
\label{sec:experiments and results}
We perform a variety of experiments and report the results in this section. The properties of ADT are studied by assessing its performance on three datasets and comparing it to benchmarks (Section \ref{sec:overall performance}), and analyzing the effect of parameters (Section \ref{sec:effect of parameters}). The thresholding performance of ADT is further validated in a feasibility study (Section \ref{sec:feasibility study}). Precision, recall, and F1 score were used to evaluate anomaly detection performance by comparing the detection result with the ground truth. All experiments were run on a machine with an 11th Gen Intel(R) Core(TM) i7-1165G7 @ 2.80GHz CPU.
\subsection{Overall Performance}
\label{sec:overall performance}
We compare ADT with two benchmarked thresholding methods, i.e. Static and DSPOT, on three real-world datasets. For each dataset, the thresholding methods can generate the optimal thresholds based on the anomaly scores produced by a well-trained AE. The overall performance of the methods are detailed in Table ~\ref{tab:overall performance}.

According to Table \ref{tab:overall performance}, ADT greatly outperforms the benchmarks, achieving the highest F1 scores on all datasets. The F1 on Yahoo is about 0.95 and that on SWaT and WADI almost reaches 1.0. In comparison, the Static and DSPOT methods obtain extremely poor results on Yahoo and WADI, i.e. the second-best F1 on Yahoo is only 0.12 from DSPOT and the second-best F1 on WADI is 0.35 from Static. The Static and DSPOT methods perform relatively better on SWaT with F1 scores of around 0.77, however, still much worse than ADT. 

Compared with Static, DSPOT shows limited improvement in the F1 on Yahoo and SWaT, whereas even worse performance on WADI. This validates that DSPOT cannot always guarantee good results and its advantage over the Static method is uncertain. 

Apart from the performance comparison through the detection over the whole dataset, we further test the robustness of the models over 10 evenly-divided distinct subsets of the dataset and report the mean and variance. According to \cite{usbeck2015gerbil}, in some special cases, if the numbers of $TP$, $FP$, and $FN$ are all 0, the precision, recall, and F1 will be considered as 1; if the number of $TP$ is 0 and one of the numbers of $FP$ and $FN$ is not 0, the precision, recall, and F1 will be determined as 0. The two-tailed Wilcoxon test \cite{demvsar2006statistical} with a $p$ value of 0.05 was conducted over each dataset to determine that the difference between our method and the benchmarks is statistically significant. The experimental results are illustrated in Table \ref{tab:performance subesets}. It demonstrates that ADT once again achieves the best performance with the highest means of evaluation metrics and the lowest variances over all datasets. On the contrary, the Static and DSPOT methods are not robust enough because their results on SWaT decrease greatly compared with the corresponding results in Table \ref{tab:overall performance}.

\begin{table*}[t]
\caption{Performance comparison on each dataset. The optimal static thresholding method (Static) and the EVT-based statistical dynamic thresholding method (DSPOT) are compared with our agent-based dynamic thresholding approach (ADT). Precision, Recall, and F1 score are reported.}
\label{tab:overall performance}
\begin{tabular}{cccccccccc} \toprule
\multirow{3}{*}{\textit{Methods}} & \multicolumn{3}{c}{\textit{Yahoo}}   & \multicolumn{3}{c}{\textit{SWaT}}    & \multicolumn{3}{c}{\textit{WADI}}    \\ \cline{2-10} 
                         & $P$       & $R$       & $F1$      & $P$       & $R$       & $F1$      & $P$       & $R$       & $F1$      \\ \midrule
Static                       & 0.02543 & 0.20611 & 0.04528 & 0.98188 & 0.63257 & 0.76943 & 0.51784 & 0.25890 & 0.34521 \\ 
DSPOT                    & 0.09781 & 0.14796 & 0.11777 & 0.98347 & 0.63913 & 0.77476 & 0.06680 & 0.65170 & 0.12117 \\ 
ADT                      & 0.95206 & 0.95175 & \textbf{0.95191} & 0.99936 & 0.99873 & \textbf{0.99905} & 0.99870 & 0.99739 & \textbf{0.99804} \\ \bottomrule
\end{tabular}
\end{table*}

\begin{table*}[t]
\caption{Average performance ($\pm$standard deviation) on 10 distinct subsets of each dataset. $\downarrow$ indicates a significant decrease in the result compared with the corresponding result shown in Table 2.}
\label{tab:performance subesets}
\begin{tabular}{cccccccccc} \toprule
\multirow{3}{*}{\textit{Methods}} & \multicolumn{3}{c}{\textit{Yahoo}}   & \multicolumn{3}{c}{\textit{SWaT}}    & \multicolumn{3}{c}{\textit{WADI}}    \\ \cline{2-10} 
                         & $P$       & $R$       & $F1$      & $P$       & $R$       & $F1$      & $P$       & $R$       & $F1$      \\ \midrule
Static                       & 0.081(0.01) & 0.250(0.02) & 0.098(0.01) & 0.574(0.16)$\downarrow$ & 0.191(0.07)$\downarrow$ & 0.252(0.09)$\downarrow$ & 0.428(0.20) & 0.299(0.18) & 0.305(0.16) \\ 
DSPOT                    & 0.165(0.06) & 0.132(0.01) & 0.100(0.02) & 0.543(0.16)$\downarrow$ & 0.198(0.08)$\downarrow$ & 0.257(0.10)$\downarrow$ & 0.142(0.07) & 0.326(0.17) & 0.131(0.05) \\ 
ADT                      & 0.946(<0.01) & 0.945(<0.01) & \textbf{0.945(<0.01)} & 0.999(<0.01) & 0.997(<0.01) & \textbf{0.998(<0.01)} & 0.999(<0.01) & 0.999(<0.01) & \textbf{0.999(<0.01)} \\ \bottomrule
\end{tabular}
\end{table*}

\subsection{Effect of Parameters}
\label{sec:effect of parameters}
The effects of the important parameters on the performance of ADT are studied in this section. Experiments were done using the SWaT dataset.

\begin{figure}[h]
    \centering
    \includegraphics[width=1\linewidth]{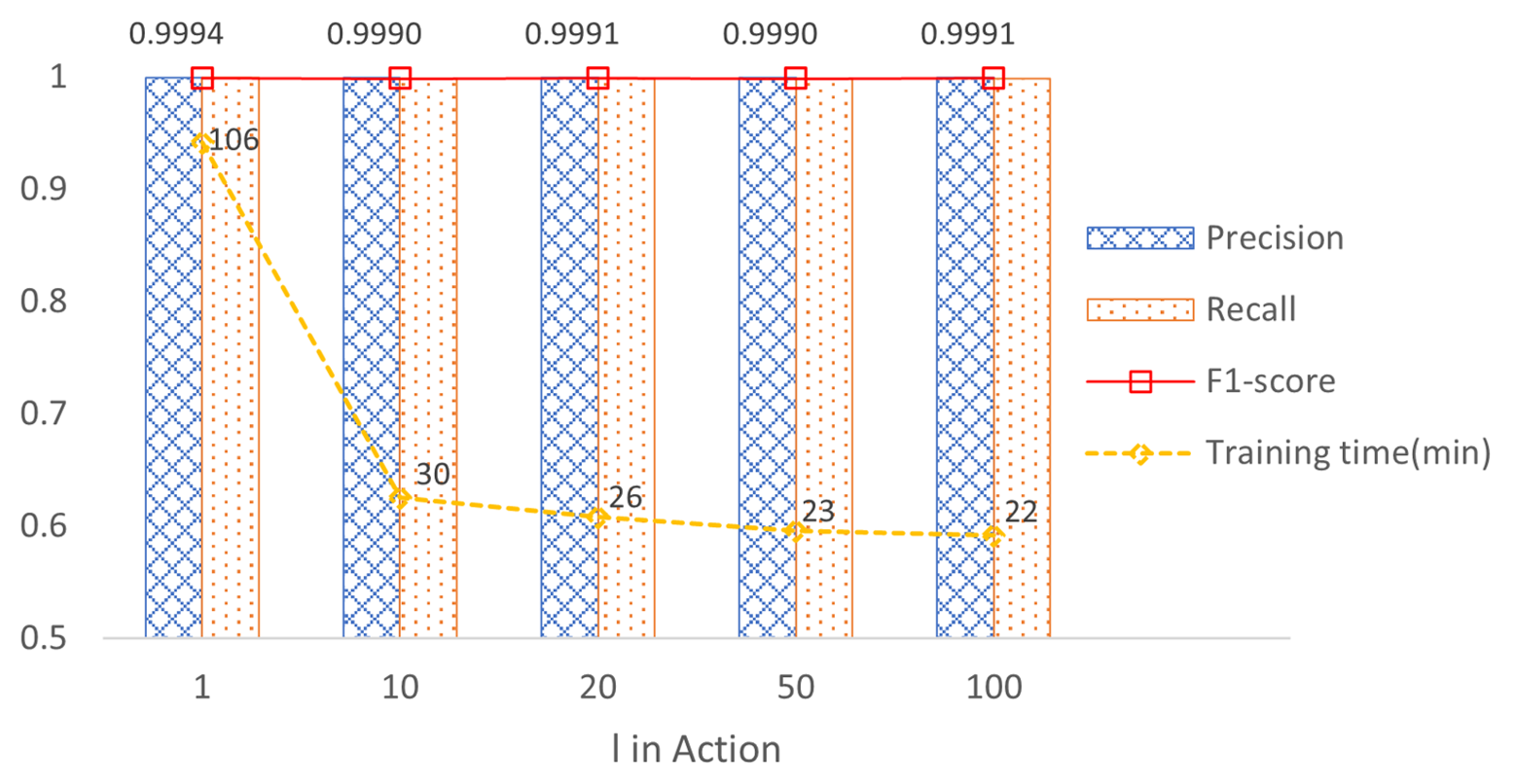}
    \caption{Effect of parameter l on Precision, Recall, F1 score, and training time}
    \label{fig:l_action}
    \Description{Effects of parameter l on Precision, Recall, F1 score, and training time.}
\end{figure}

\begin{figure*}[h]
    \centering
    \includegraphics[width=1.0\linewidth]{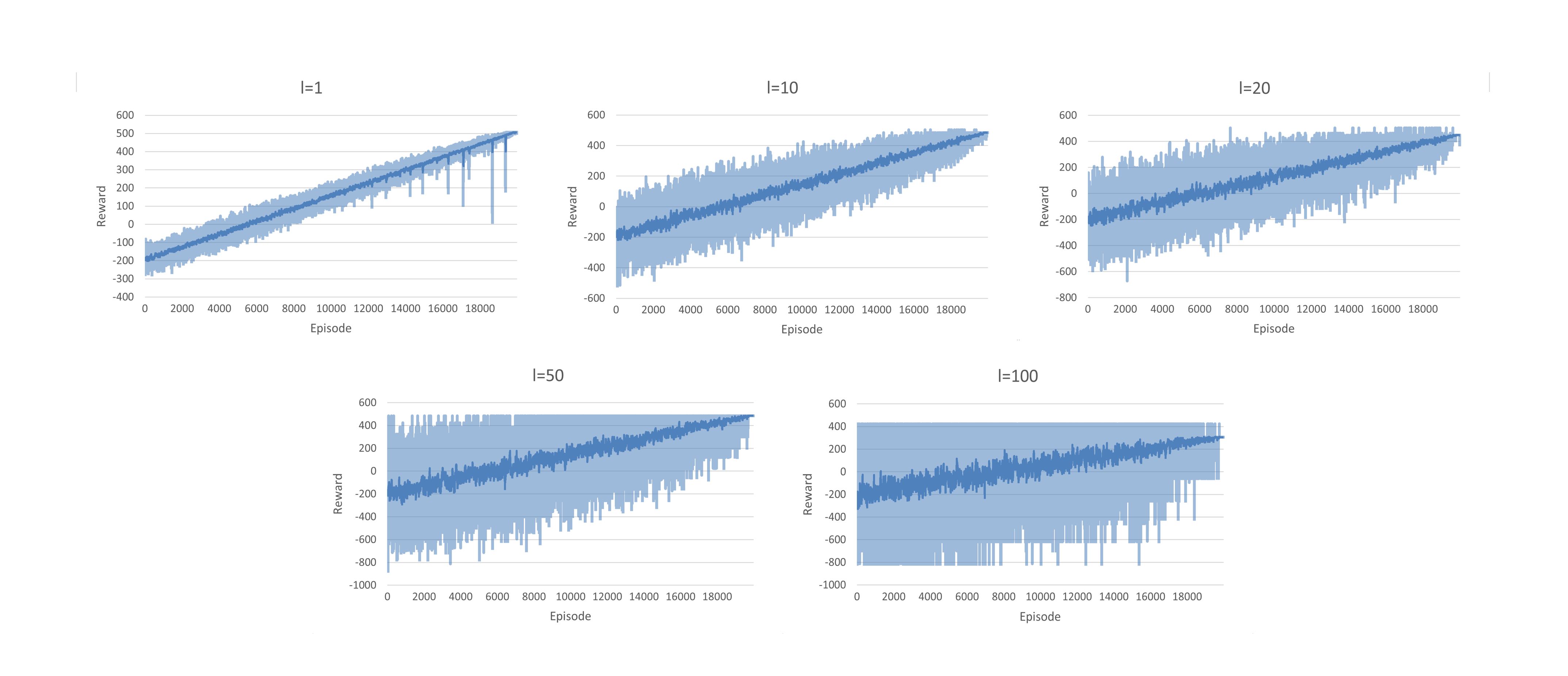}
    \caption{Effect of parameter $l\in \{1, 10, 20, 50, 100\}$ on the training stability of ADT. The darker blue line in each sub-figure depicts the moving average of every 30 episodes.}
    \label{fig:training_l}
    \Description{Effects of parameter l on training time. Generally, a smaller $l$ achieve more stable training.}
\end{figure*}

The first factor we study is how our model responds to different values of $l$ used in the agent's action selection. The DQN agent changes its action $a$ per $l$ time steps ($l\geq1$) for higher training efficiency. If $t\mathbin{\%}l=0$, the action $a_t$ is from the decayed $\epsilon$-greedy strategy; otherwise, the action is maintained as $a_t=a_{t-1}$. Figure ~\ref{fig:l_action} summarises the obtained results using 5 different values $l \in \{1, 10, 20, 50, 100\}$ on precision, recall, F1 score, and the training time. Figure ~\ref{fig:training_l} presents the results about the training stability. According to Figure ~\ref{fig:l_action}, the inference performance of ADT is relatively insensitive to $l$ because the evaluation metrics in different cases are almost the same (e.g. F1 scores are all around 0.999). The model's training time is more related to $l$, and in general a smaller $l$ causes more training time. Figure ~\ref{fig:training_l} depicts the rewards obtained during the training process, and the darker blue line in each sub-figure is the moving average of every 30 episodes for smoothing. It shows that a larger $l$ results in more instability of the training. Thus, the selection of the value of $l$ should not only account for the detection performance but also the trade-off between training efficiency and stability. 

\begin{figure}[h]
    \centering
    \includegraphics[width=1\linewidth]{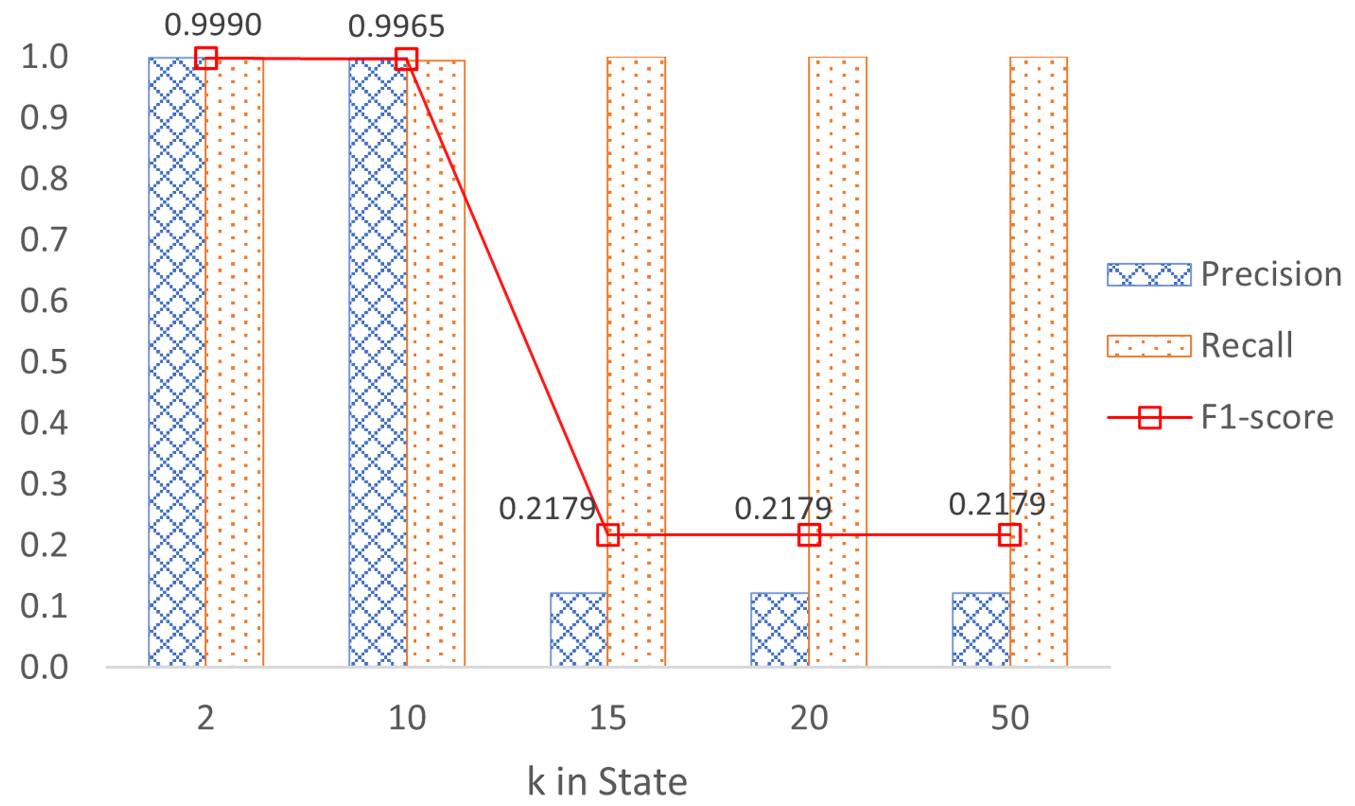}
    \caption{Effect of parameter k on Precision, Recall, and F1 score}
    \label{fig:k_state}
    \Description{Effects of parameter k on Precision, Recall, and F1 score.}
\end{figure}

The second factor we investigate is how our model responds to different values of $k$ used in the environment's state representation. At time step $t$, the state $s_{t}$ is a tuple of six (see Section ~\ref{sec:dqn-state}) where each element describes some critical feature of the previously encountered $k$ time windows. Figure ~\ref{fig:k_state} presents the obtained results for 5 different values of $k \in \{2, 10, 15, 20, 50\}$. We observe that with the increase of $k$, the recall always keeps high, while the precision and F1 stay high at the beginning but then decrease until reaching the minimum value, thus leading to poor performance at the end. It implies that when $k$ exceeds a certain value, it causes a dramatic increase in the number of $FPs$ which highly limits the detection performance. Therefore, a suitable $k$ value should be relatively small.

The last parameters we study are $\alpha$ and $\beta$ which are used for the weighted average of $TP-FN-FP$ and $TN$ in the reward function (see Equation \ref{eqn:new reward}). Empirically, the selection of $\alpha$ and $\beta$ is related to several factors such as the dataset, the choice of $l$, etc. For the SWaT dataset, Table ~\ref{table:alpha_beta} reports the effect of $\alpha$, $\beta$ in F1 score with different values of $l\in \{10, 50, 100\}$. It indicates that when $l$ changes, the obtained results for different values of $\alpha$ and $\beta$ will change. For example, $\alpha=1$ and $\beta=0$ leads to a high F1 of 0.999 when l=10, whereas low F1 of 0.455 when l=50 and l=100. We observe that among all the options, $\alpha=0.9$, $\beta=0.1$ and $\alpha=0.5$, $\beta=0.5$ both produce relatively good results regardless of the value of $l$. 

\begin{table}[t]
  \caption{Effect of $\alpha$ and $\beta$ on F1 score with $l\in \{10, 50, 100\}$}
  \label{table:alpha_beta}
  \begin{tabular}{ccccc}\toprule
    $\alpha$ & $\beta$ & $l=10$ & $l=50$ & $l=100$ \\ \midrule
    1    & 0    & 0.99936 & 0.45512  & 0.45512 \\
   0.9   & 0.1  & 0.99905 & 0.99905  & 0.99907 \\
   0.5   & 0.5  & 0.99905 & 0.99905  & 0.99905 \\
   0.1   & 0.9  & 0.99905 & 0.95373  & 0.87527 \\
   0     & 1    & 0.88679 &  0.05085 & 0.05989 \\ \bottomrule
  \end{tabular}
\end{table}
\subsection{Feasibility Study}
\label{sec:feasibility study}
\begin{figure*}[h]
    \centering
    \includegraphics[width=1.0\linewidth]{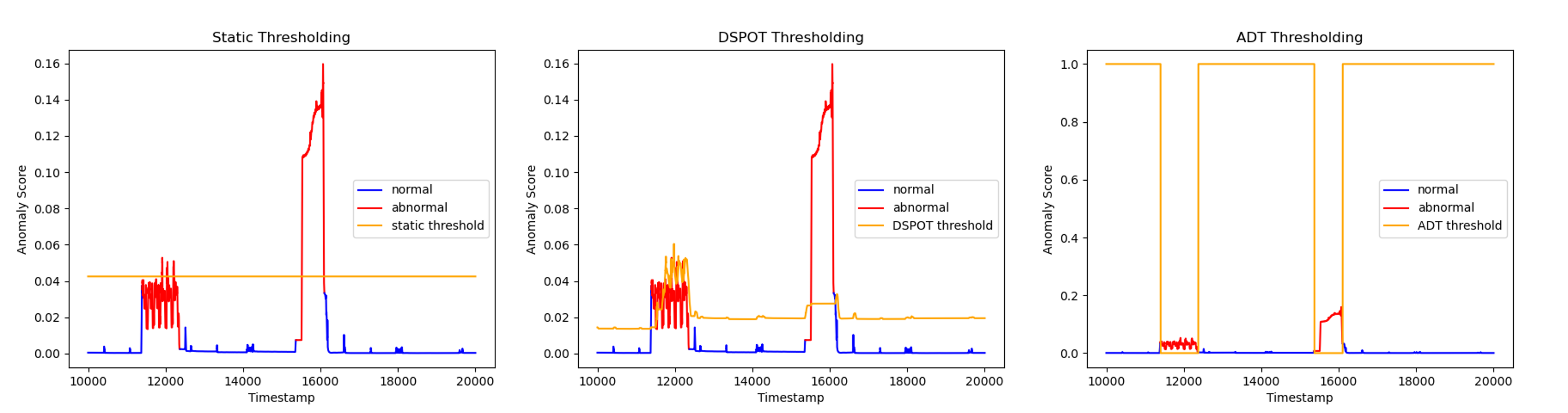}
    \caption{Thresholding performance on a time series from the SWaT dataset. The $y$-axis values represent the calculated anomaly scores for the time series. According to the ground truth, we color the abnormal instances occurring at $11399\le t\le12373$ and $15369\le t\le16100$ in red and the normal instances in blue. The orange line represents the threshold values generated by the thresholding method.}
    \label{fig:compare_thresholds}
    \Description{Examples of detection results of a time series using different thresholding method.}
\end{figure*}

We perform anomaly detection with all thresholding methods on a time series extracted from the SWaT dataset and compare the thresholding performance in Figure \ref{fig:compare_thresholds}. To obtain an accurate prediction for a time window, the generated threshold value should be lower than the anomaly score if the ground truth of the window is abnormal, and vice versa.

Figure \ref{fig:compare_thresholds} reveals that both the Static and DSPOT methods can give proper thresholds in normal or most of the normal segments. However, their generated thresholds in abnormal segments (especially the first abnormal segment) are likely to be improper, which leads to many false negatives. This further validates that DSPOT has difficulty providing the most suitable continuous thresholds in some cases. In contrast, ADT can correctly transform the threshold to the active mode ($\delta=0$) in abnormal segments and the passive mode ($\delta=1$) in normal segments, showing the best thresholding performance in anomaly detection. The classification accuracy with ADT, i.e. the percentage of $TPs$ and $TNs$ out of all $TPs$, $TNs$, $FPs$, and $FNs$ reaches 0.9994.

\section{Conclusion and Future Work}
\label{sec:conclusion}
Setting appropriate thresholds in anomaly detection is critical and challenging. Conventional thresholding methods, e.g. static or manually-defined thresholds, are no longer useful on complex data. This paper models thresholding in anomaly detection as a MDP and presents an agent-based dynamic thresholding framework (ADT) based on DQN. The proposed method can be integrated with many systems that require dynamic thresholding. ADT can on that basis utilize the abnormal-related measurements and provide adaptive optimal thresholding control. 

Our method can be used as an anomaly detector, but more prone for dynamic thresholding. The key points are that it is very data-efficient, and it transforms thresholding in anomaly detection to the passive mode and the active mode. Through a variety of experiments on three real-world datasets and comparison with two benchmarks, our agent-based method shows outstanding thresholding capability, stability, and robustness, leading eventually to significantly improved anomaly detection performance. 

One limitation of our method is that the detection performance may be compromised when the dataset is characterized by some isolated point anomalies, which is worth studying in the future. Furthermore, we intend to explore continuous thresholding and multi-objective reward function within the ADT framework. From the practical perspective, we would like to apply ADT in more complex systems and scenarios.




\begin{acks}
This work is funded by the College of Science and Engineering Postgraduate Research Scholarship of the University of Galway.
\end{acks}



\bibliographystyle{ACM-Reference-Format} 
\bibliography{sample}


\end{document}